\documentclass{article}

\usepackage{microtype}
\usepackage{graphicx}
\usepackage{subfigure}
\usepackage{booktabs} 

\usepackage{hyperref}

\usepackage{bm}

\usepackage{amsthm}
\newtheorem{prop}{Proposition}
\newtheorem{Theorem}{Theorem}
\newtheorem{Remark}{Remark}
\newtheorem{Corollary}{Corollary}

\usepackage{tabularx}
\newcolumntype{L}{>{\raggedright\arraybackslash}X}
\newcolumntype{C}{>{\centering\arraybackslash}X}

\usepackage[accepted]{icml2025}

\usepackage{amsmath}
\usepackage{amssymb}
\usepackage{mathtools}
\usepackage{amsthm}

\usepackage{booktabs} 
\usepackage{multirow} 
\usepackage{array}    

\usepackage[capitalize,noabbrev]{cleveref}

\theoremstyle{plain}

\theoremstyle{definition}

\theoremstyle{remark}

\usepackage[textsize=tiny]{todonotes}

\icmltitlerunning{Kernel VICReg for Self-Supervised Learning in Reproducing Kernel Hilbert Space}

\begin{document}

\AddToShipoutPictureBG*{%
  \AtPageUpperLeft{%
    \setlength\unitlength{1in}%
    \hspace*{\dimexpr0.5\paperwidth\relax}
    \makebox(0,-0.75)[c]{\normalsize {\color{black} Published in Big Data and Cognitive Computing, 2026, volume 10, issue 3, } \url{https://doi.org/10.3390/bdcc10030078}}
    }}

\twocolumn[
\icmltitle{Kernel VICReg for Self-Supervised Learning \\in Reproducing Kernel Hilbert Space}



\icmlsetsymbol{equal}{*}

\begin{icmlauthorlist}
\icmlauthor{M. Hadi Sepanj}{equal,yyy}
\icmlauthor{Benyamin Ghojogh}{equal,aaa}
\icmlauthor{Saed Moradi}{yyy}
\icmlauthor{Paul Fieguth}{yyy}
\end{icmlauthorlist}

\icmlaffiliation{yyy}{Vision and Image Processing Group, Systems Design Engineering, University of Waterloo, ON, Canada.}
\icmlaffiliation{aaa}{Electrical and Computer Engineering, University of Waterloo, ON Canada.}

\icmlcorrespondingauthor{M.Hadi Sepanj}{mhsepanj@uwaterloo.ca}
\icmlcorrespondingauthor{Benyamin Ghojogh}{bghojogh@uwaterloo.ca}
\icmlcorrespondingauthor{Saed Moradi}{saed.moradi@uwaterloo.ca}
\icmlcorrespondingauthor{Paul Fieguth}{paul.fieguth@uwaterloo.ca}

\icmlkeywords{Machine Learning, ICML}

\vskip 0.3in
]



\printAffiliationsAndNotice{\icmlEqualContribution} 

\begin{abstract}
Self-supervised learning (SSL) has emerged as a powerful paradigm for representation learning by optimizing geometric objectives, such as invariance to augmentations, variance preservation, and feature decorrelation, without requiring labels. However, most existing methods operate in Euclidean space, limiting their ability to capture nonlinear dependencies and geometric structures. In this work, we propose Kernel VICReg, a novel self-supervised learning framework that pulls the VICReg objective into a Reproducing Kernel Hilbert Space (RKHS). By kernelizing each term of the loss, variance, invariance, and covariance, we obtain a general formulation that operates on double-centered kernel matrices and Hilbert--Schmidt norms, enabling nonlinear feature learning without explicit mappings. We demonstrate that Kernel VICReg mitigates the risk of representational collapse under challenging conditions and improves performance on datasets exhibiting nonlinear structure or limited sample regimes. Empirical evaluations across MNIST, CIFAR-10, STL-10, TinyImageNet, and ImageNet100 show consistent gains over Euclidean VICReg, with particularly strong improvements on datasets where nonlinear structures are prominent. UMAP visualizations are provided only as a qualitative illustration of embedding geometry and are not used as a calibration or statistical validation. Our results suggest that kernelizing SSL objectives is a promising direction for bridging classical kernel methods with modern representation learning.
\end{abstract}


\section{Introduction}
Self-supervised learning (SSL) has emerged as a dominant paradigm for representation learning by leveraging the underlying structure of data without the need for human-annotated labels \cite{chen2020simple,bardes2022vicreg}. Methods such as SimCLR~\cite{chen2020simple}, BYOL~\cite{grill2020bootstrap}, VICReg~\cite{bardes2022vicreg}, and Barlow Twins~\cite{zbontar2021barlow} have demonstrated remarkable performance by enforcing objectives such as invariance to augmentations, feature decorrelation, and variance preservation. However, relying on Euclidean representations in standard self-supervised learning objectives often assumes a relatively simple geometric structure in the latent space. After multiple layers of nonlinear transformation, this assumption becomes questionable, since latent representations are likely to inhabit a highly non-linear manifold, poorly characterized by standard second-order statistics or $\ell_2$ distances. This motivates our kernelized formulation, which enables learning in an implicitly defined high-dimensional feature space that captures the underlying manifold structure.

{
Several recent works have incorporated kernels into SSL objectives, e.g., \cite{li2021self,wu2025pseudo,ni2024graph,sepanj2025self,sepanj2024aligning}. 
These methods typically replace similarity metrics or introduce kernel-based dependence criteria within contrastive or predictive frameworks.
In contrast, our approach performs a structural lifting of VICReg itself: the variance and covariance penalties are rederived from the covariance operator in RKHS, rather than modified heuristically. 
This distinction is important, as it preserves the collapse-prevention mechanism of VICReg while redefining its geometry.
}


Kernel methods, long celebrated for their ability to implicitly map data into high-dimensional feature spaces via the kernel trick~\cite{scholkopf2002learning}, offer a compelling avenue to address this limitation. In supervised settings, the transformation from linear to nonlinear models via kernelization (exemplified by the transition from linear SVM to kernel SVM) has been foundational in classical machine learning. Inspired by this paradigm, we ask: \textit{can core SSL losses be systematically lifted into a Reproducing Kernel Hilbert Space?} 
 Here, we answer this by showing how one can replace Euclidean-space objectives with their RKHS counterparts, exemplified through a kernelized version of VICReg.

We propose \emph{Kernel-VICReg}, which computes invariance, variance, and covariance entirely in RKHS via double-centered kernels and Hilbert--Schmidt norms, while we focus on VICReg as a concrete example (and note that Barlow Twins admits an analogous kernelization), the same RKHS-lifting could be extended to contrastive frameworks like SimCLR or predictive ones like BYOL with suitable cross-kernel formulations.

{ Experiments across multiple datasets show that Kernel-VICReg can yield more effective representations than its Euclidean counterpart under our evaluation protocol and can improve stability in settings where Euclidean VICReg collapses. These results suggest that integrating kernel methods into SSL is a promising direction.}

{
\subsection{Related Work: Kernel Methods in Self-Supervised Learning}
While several recent SSL methods incorporate kernel functions, they fundamentally differ from Kernel VICReg in their application and scope. Existing approaches generally fall into the following categories:
\begin{itemize}
    \item Use of Nonlinear Dependence in SSL: Some methods such as \cite{li2021self,sepanj2025self} use Hilbert--Schmidt Independence Criterion (HSIC) \cite{gretton2005measuring} to model nonlinear dependence of samples or features in RKHS.
    \item Kernels as Regularizers: Some methods, such as \cite{sepanj2024aligning}, integrate kernel tools, such as Maximum Mean Discrepancy (MMD) \cite{gretton2006kernel}, into existing SSL losses to align feature distributions. In these cases, the kernel is used as a supplementary distance metric, but the underlying architecture remains Euclidean.
    \item Implicit Kernel Analysis: Other works, such as \cite{simon2023stepwise}, use kernels primarily as a theoretical lens to analyze the training dynamics or ``spectral bias'' of standard SSL objectives like SimCLR \cite{chen2020simple} or Barlow Twins \cite{zbontar2021barlow}.
\end{itemize}

In 
contrast, our proposed Kernel VICReg represents a systematic lifting of an entire loss function of an existing SSL method, i.e., VICReg, into the RKHS. Unlike methods that use kernels for specific terms, we kernelize the variance, invariance, and covariance terms simultaneously. This allows the model to operate on double-centered kernel matrices and Hilbert--Schmidt norms, capturing nonlinear dependencies across the entire loss function rather than just regularizing a single component. To the best of our knowledge, this is the first work to provide a complete kernelized derivation of the VICReg framework.
}

{
\subsection{Positioning and Novelty}

Recent works have explored the use of kernels in self-supervised learning, including kernel-based contrastive objectives and kernelized predictive frameworks. However, these methods typically replace similarity measures or dependence criteria within existing objectives (e.g., kernelized contrastive alignment or kernel dependence maximization), rather than systematically lifting the entire regularization structure of a non-contrastive SSL method into RKHS.

Our contribution is structurally different. 
We show that all three components of VICReg---invariance, variance preservation, 
and covariance decorrelation---admit principled RKHS counterparts derived from 
the covariance operator in Hilbert space. 
This yields a unified formulation in which:
\begin{enumerate}
    \item invariance is expressed via cross-kernel trace distances, 
    \item variance preservation is tied to eigenvalues of the centered kernel matrix, and  
    \item covariance decorrelation becomes a Hilbert--Schmidt norm penalty.
\end{enumerate}

To 
our knowledge, a full operator-level lifting of VICReg into RKHS using covariance operators and double-centered Gram matrices has not been previously derived.
Importantly, our formulation is not a kernel substitution in the similarity function but a redefinition of the geometry in which the SSL objective is defined.

The proposed Kernel VICReg framework provides a more robust geometric constraint in the RKHS, while standard Euclidean methods may suffer from dimensional collapse when the projection head is not sufficiently wide. Our kernelized approach enhances robustness to representation collapse by leveraging the infinite-dimensional nature of the Hilbert space to maintain feature variance.
}

\section{Review of VICReg}

\subsection{Data and Network Settings}

The neural network is composed of an encoder network and an expander network. 
Let 
$\mathcal{X} := \{\bm{x}_1, \bm{x}_2, \dots, \bm{x}_n\}$ be the dataset where $\bm{x}_i \in \mathbb{R}^d$ and $d$ is the dimensionality of data. Every batch of data, with size $b$, contains $\mathcal{X}_b := \{\bm{x}_1, \bm{x}_2, \dots, \bm{x}_b\}$ and $\mathcal{X}'_b := \{\bm{x}'_1, \bm{x}'_2, \dots, \bm{x}'_b\}$ where $\bm{x}_i$ and $\bm{x}'_i$ correspond to each other; e.g., they are augmentations of a common underlying image. They pass through an encoder network to output the $q$-dimensional latent embeddings $\{\bm{y}_i\}_{i=1}^b$ and $\{\bm{y}'_i\}_{i=1}^b$ where $\bm{y}_i, \bm{y}'_i \in \mathbb{R}^q$. The latent embeddings pass through the expander network to obtain the $p$-dimensional output embeddings $\{\bm{z}_i\}_{i=1}^b$ and $\{\bm{z}'_i\}_{i=1}^b$ where $\bm{z}_i, \bm{z}'_i \in \mathbb{R}^p$ where $p > q$. 


\subsection{VICReg Loss Function}

VICReg~\cite{bardes2022vicreg} is a non-contrastive SSL method that simultaneously enforces three complementary properties on a pair of views \(\bm{x}\) and \(\bm{x}'\) of the same sample:  
\begin{enumerate}
  \item Invariance: embeddings of corresponding views should be close;
  \item Variance preservation: each dimension of the embedding space should maintain sufficient spread to avoid collapse;
  \item Covariance decorrelation: different embedding dimensions should be de-correlated with encourage richness.
\end{enumerate}

\textls[-15]{The 
VICReg loss function contains three terms reflecting the three preceding properties:}

The \textit{invariance loss} is defined as: 

\begin{align}
\mathcal{L}_{\text{inv}}(\bm{x}, \bm{x}') = \frac{1}{b}\sum_{i=1}^b \big\lVert \bm{z}_i - \bm{z}'_i \big\rVert^2_2,
\end{align}
which encourages paired views to have similar representations.

The \textit{variance loss} seeks to ensure that no dimension collapses to zero variance, and is defined as
\begin{align}
  \mathcal{L}_{\text{var}}(\bm{x}) = \frac{1}{p}
    \sum_{j=1}^p 
    \bigl[\gamma - \sigma_j \bigr]_{+}^{2},
\end{align}
for threshold \(\gamma>0\) and where \([.]_+ = \max(.,0)\).
The sample standard deviation is found as:
\begin{align}
\sigma_j = \sqrt{\textbf{Var}(\{z_{i,j}\}_{i=1}^b) + \epsilon},
\end{align}
where parameter \(\epsilon>0\) prevents degeneracy,
and $z_{ij}$ denotes the $j$-th dimension of $\bm{z}_i$.

For the \textit{covariance loss},
let 
\begin{align}
    \bm{C} = 1/(b-1) \widetilde{\bm{Z}}^\top \widetilde{\bm{Z}} \in \mathbb{R}^{p \times p}, 
\end{align}
 be the empirical covariance of the zero-centered embeddings \(\widetilde{\bm{Z}}\in\mathbb{R}^{b\times p}\).  VICReg penalizes off-diagonal correlations:
\begin{align}\label{equation_original_vicreg_covariance_loss}
  \mathcal{L}_{\text{cov}}(\bm{x})
  = \frac{1}{p}\sum_{j=1}^p \sum_{k=1, k \neq j}^p \bm{C}_{jk}^{2},
\end{align}
where $\bm{C}_{jk}$ denotes the $(j,k)$-th element of the matrix $\bm{C}$.
Promoting a diagonal covariance matrix encourages different features to capture distinct aspects of the data.

The \textit{Overall VICReg Loss}
then is a weighted summation of the preceding terms:
\begin{equation}
\begin{aligned}
  \mathcal{L}_{\text{VICReg}}
  =\, &\alpha \mathcal{L}_{\text{inv}}(\bm{x}, \bm{x}') + \beta \big( \mathcal{L}_{\text{var}}(\bm{x}) + \mathcal{L}_{\text{var}}(\bm{x}') \big) \\
    &+ \zeta \big( \mathcal{L}_{\text{cov}}(\bm{x}) + \mathcal{L}_{\text{cov}}(\bm{x}') \big),
\end{aligned}
\end{equation}
with hyperparameters \(\alpha,\beta,\zeta>0\).  VICReg thus avoids the need for negative samples or momentum encoders by combining these three regularizers in Euclidean space, laying the groundwork for our kernelized extension in the next section.

\section{Kernel VICReg}

Standard VICReg operates in  Euclidean space by enforcing variance preservation, decorrelating feature dimensions, and ensuring consistency between views of the same sample. Our proposed Kernel VICReg extends these principles to the Reproducing Kernel Hilbert Space (RKHS), allowing for non-linear representations without explicit feature mappings. We rigorously derive the kernelized counterparts of VICReg's variance, covariance, and invariance losses.

\subsection{Covariance in RKHS}

Before deriving the kernelized VICReg loss components, we establish the fundamental result that the covariance operator in RKHS is proportional to the double-centered kernel matrix. 
The formulations derived and explained in this section will be used in kernelizing the loss terms of VICReg. 

Let $\phi(\bm{x})$ be the implicit feature mapping in RKHS $\mathcal{H}$, where $\phi(\cdot)$ is the pulling function into RKHS. The kernel function in the RKHS $\mathcal{H}$ is defined as:
\begin{align}
k(\bm{x}_i, \bm{x}_j) = \langle \phi(\bm{x}_i), \phi(\bm{x}_j) \rangle_{\mathcal{H}},
\end{align}
where $\langle \cdot, \cdot \rangle_{\mathcal{H}}$ denotes the inner product in the RKHS $\mathcal{H}$.

The covariance operator in RKHS is defined as \cite{scholkopf1997kernel,scholkopf1998nonlinear}:
\begin{equation}\label{equation_covariance_in_RKHS}
\mathbb{R}^{t \times t} \ni \bm{C}_{\phi}(\bm{x}) = \frac{1}{b} \sum_{i=1}^{b} \left(\phi(\bm{x}_i) - \bar{\phi}\right) \left(\phi(\bm{x}_i) - \bar{\phi}\right)^\top,
\end{equation}
where $t$ is the dimensionality of the RKHS, $\phi(\bm{x}_i) - \bar{\phi}$ is the centered feature map of the batch, and $\bar{\phi} = \frac{1}{b} \sum_{j=1}^{b} \phi(\bm{x}_j)$ is the mean feature vector of the batch in RKHS.
If we let
\begin{align}\label{equation_Phi_matrix}
\bm{\Phi} := [\phi(x_1), \dots, \phi(x_b)] \in \mathbb{R}^{t \times b},
\end{align}
then Equation (\ref{equation_covariance_in_RKHS}) can be restated in matrix form:
\begin{equation}\label{equation_covariance_in_RKHS_matrix_form}
\mathbb{R}^{t \times t} \ni \bm{C}_{\phi}(\bm{x}) = \frac{1}{b} (\bm{\Phi H}) (\bm{\Phi H})^\top,
\end{equation}
where $\bm{H} := \bm{I}_b - \frac{1}{b} \bm{1}_b \bm{1}_b^\top \in \mathbb{R}^{b \times b}$ is the centering matrix. 

Using the kernel trick, we define kernel matrix $\bm{K} \in \mathbb{R}^{b \times b}$ whose $(i,j)$-th element is:
\begin{equation}
    \bm{K}(\bm{x})[i, j] = k(\bm{x}_i, \bm{x}_j) = \langle \phi(\bm{x}_i), \phi(\bm{x}_j) \rangle_{\mathcal{H}}.
\end{equation}
From 
Equation (\ref{equation_Phi_matrix}), the kernel matrix can be stated as:
\begin{align}\label{equation_kernel_with_P}
\bm{K}(x) = \bm{\Phi}^\top \bm{\Phi}.
\end{align}
The double-centered kernel matrix is:
\begin{equation}
\begin{aligned}\label{equation_double_centered_kernel}
&\widehat{\bm{K}}(\bm{x}) := \bm{H} \bm{K}(\bm{x}) \bm{H} \overset{(\ref{equation_kernel_with_P})}{=} \bm{H} \bm{\Phi}^\top \bm{\Phi} \bm{H} \overset{(s)}{=} (\bm{\Phi H})^\top (\bm{\Phi H}),
\end{aligned}
\end{equation}
where step $(s)$ considers that the centering matrix $\bm{H}$ is symmetric. 
The double-centered kernel matrix will be used in the following. 

The following explains the relation of the covariance operator and double-centered kernel in RKHS.
The squared Hilbert--Schmidt norm of the covariance matrix in RKHS is:
\begin{align}
\|\bm{C}_{\phi}(\bm{x})\|_{\text{HS}}^2 &= \textbf{tr}(\bm{C}_{\phi}(\bm{x})^\top \bm{C}_{\phi}(\bm{x})) \nonumber \\
&\overset{(\ref{equation_covariance_in_RKHS_matrix_form})}{=} \frac{1}{b^2} \textbf{tr}((\bm{\Phi H}) (\bm{\Phi H})^\top (\bm{\Phi H}) (\bm{\Phi H})^\top) \nonumber \\
&\overset{(a)}{=} \frac{1}{b^2} \textbf{tr}((\bm{\Phi H})^\top (\bm{\Phi H}) (\bm{\Phi H})^\top (\bm{\Phi H})) \nonumber \\
&\overset{(\ref{equation_double_centered_kernel})}{=} \frac{1}{b^2} \textbf{tr}(\widehat{\bm{K}}(\bm{x}) \widehat{\bm{K}}(\bm{x})) \nonumber \\
&= \frac{1}{b^2} \textbf{tr}(\widehat{\bm{K}}(\bm{x})^2), \label{equation_squared_norm_covariance_RKHS}
\end{align}
where $\textbf{tr}(\cdot)$ denotes the trace of matrix and step $(a)$ follows from the cyclic property of the trace operator.
For easier computation in computer, it is possible to restate Equation (\ref{equation_squared_norm_covariance_RKHS}) using the Frobenius norm:
\begin{align}
\|\bm{C}_{\phi}(\bm{x})\|_{\text{HS}}^2 &= \frac{1}{b^2} \textbf{tr}(\widehat{\bm{K}}(\bm{x})^2) = \frac{1}{b^2} \sum_{i \neq j} [\widehat{\bm{K}}(\bm{x})]_{i,j}^2 \nonumber \\
&= \frac{1}{b^2} \Big( \|\widehat{\bm{K}}(\bm{x})\|_F^2 - \sum_{i=1}^b [\widehat{\bm{K}}(\bm{x})]_{i,i}^2 \Big). \label{equation_squared_norm_covariance_RKHS_2}
\end{align}

From Equation (\ref{equation_squared_norm_covariance_RKHS}), the covariance operator in RKHS is proportional to the centered kernel matrix:
\begin{equation}\label{equation_C_proportional_K}
    \bm{C}_{\phi}(\bm{x}) \propto \frac{1}{b} \widehat{\bm{K}}(\bm{x}).
\end{equation}
This key result enables the kernelization of VICReg's variance and covariance regularization terms, as discussed next.

\subsection{Kernelized Variance Regularization}

The variance regularization term in VICReg prevents representation collapse by ensuring that the variance along each feature dimension remains sufficiently large, i.e., above a threshold $\gamma$. In RKHS, the variance of feature dimensions corresponds to the eigenvalues of $\bm{C}_{\phi}(\bm{x})$. Since $\bm{C}_{\phi}(\bm{x})$ is proportional to $\widehat{\bm{K}}(\bm{x})$ according to Equation (\ref{equation_C_proportional_K}), we define the kernelized variance loss as:
\begin{equation}\label{equation_kernelized_variance_term}
    \mathcal{L}_{\text{var}}(\bm{x}) = \frac{1}{b} \sum_{i=1}^{b} \Big( \Big[\gamma - \sqrt{\frac{\lambda_i}{b} + \epsilon}\Big]_+ \Big)^2,
\end{equation}
where $[\cdot]_+ := \max(\cdot, 0)$ is the standard Hinge loss, $\{\lambda_i\}_{i=1}^b$ are the eigenvalues of $\widehat{\bm{K}}(\bm{x})$, $\gamma$ is a threshold for the minimum desired standard deviation, and $\epsilon$ is a small positive number preventing numerical instabilities. 
The proof of why $\lambda_i / b$ is understood as a variance in RKHS will be developed in Section \ref{section_relation_with_kernel_pca}.

It is noteworthy that computing the eigenvalues here is not concerning in terms of time of computation because the double-centered kernel is $b \times b$ where $b$ is the batch size, which is usually not a very large number. 

\subsection{Kernelized Covariance Regularization}

To prevent redundancy in representations, VICReg penalizes off-diagonal elements of the covariance matrix (see Equation (\ref{equation_original_vicreg_covariance_loss})). Building on Equation (\ref{equation_squared_norm_covariance_RKHS_2}), the kernelized covariance loss can be defined as
\begin{align}\label{equation_kernel_vicreg_covariance_loss}
\mathcal{L}_{\text{cov}}(\bm{x}) &= \|\bm{C}_{\phi}(\bm{x})\|_{\text{HS}} \nonumber \\
&= \frac{1}{b} \sqrt{\Big( \|\widehat{\bm{K}}(\bm{x})\|_F^2 - \sum_{i=1}^b [\widehat{\bm{K}}(\bm{x})]_{i,i}^2 \Big)}.
\end{align} 
Because of the direct relation between covariance and correlation, this regularization enforces decorrelation between features in RKHS.


{
The choice of using the Hilbert--Schmidt norm $\|\bm{C}_{\phi}(\bm{x})\|_{\text{HS}}$ in Equation (\ref{equation_kernel_vicreg_covariance_loss}) rather than its square $\|\bm{C}_{\phi}(\bm{x})\|_{\text{HS}}^2$ (as initially suggested by the expansion in Equation (\ref{equation_squared_norm_covariance_RKHS_2})) is a deliberate design choice aimed at improving optimization stability. Mathematically, if $\lambda_i$ are the singular values of the covariance operator, the squared norm $\|\bm{C}_{\phi}(\bm{x})\|_{\text{HS}}^2 = \sum \lambda_i^2$ penalizes large correlations quadratically, which can lead to vanishing gradients for smaller correlation values during the late stages of training. By using the square root form (the norm itself), the gradient magnitude remains more consistent:
\begin{equation}
    \frac{\partial \|\bm{C}_{\phi}(\bm{x})\|_{\text{HS}}}{\partial \theta} = \frac{1}{2 \|\bm{C}_{\phi}(\bm{x})\|_{\text{HS}}} \frac{\partial \|\bm{C}_{\phi}(\bm{x})\|_{\text{HS}}^2}{\partial \theta}.
\end{equation}
Empirically, we observed that this formulation provides a more balanced optimization landscape, preventing the covariance term from being dominated by a few large off-diagonal correlations and ensuring that all dimensions of the RKHS embedding are decorrelated effectively. This is consistent with recent findings in SSL literature suggesting that normalization of loss terms can lead to smoother convergence and better numerical stability in high-dimensional feature spaces.
}

\subsection{Kernelized Invariance Term}

The invariance term of the loss function minimizes the mean squared error between corresponding samples $\bm{x}$ and $\bm{x}'$, i.e., different views of the same sample. 
Consider the following $(b \times b)$ kernel matrices:
\begin{align*}
&\bm{K}(\bm{x}, \bm{x})[i, j] = k(\bm{x}_i, \bm{x}_j) = \langle \phi(\bm{x}_i), \phi(\bm{x}_j) \rangle_{\mathcal{H}}, \\
&\bm{K}(\bm{x}', \bm{x}')[i, j] = k(\bm{x}'_i, \bm{x}'_j) = \langle \phi(\bm{x}'_i), \phi(\bm{x}'_j) \rangle_{\mathcal{H}}, \\
&\bm{K}(\bm{x}, \bm{x}')[i, j] = k(\bm{x}_i, \bm{x}'_j) = \langle \phi(\bm{x}_i), \phi(\bm{x}'_j) \rangle_{\mathcal{H}}.
\end{align*}

Given kernel matrices $\bm{K}(\bm{x}, \bm{x})$ and $\bm{K}(\bm{x}', \bm{x}')$ for two augmented views and their cross-kernel matrix $\bm{K}(\bm{x}, \bm{x}')$, the distance of the views in RKHS is defined as \cite{scholkopf2000kernel}:
\begin{equation}
    \mathcal{L}_{\text{inv}}(\bm{x}, \bm{x}') = \frac{1}{b} \textbf{tr}\big(\bm{K}(\bm{x}, \bm{x}) + \bm{K}(\bm{x}', \bm{x}') - 2\bm{K}(\bm{x}, \bm{x}')\big).
\end{equation}
This loss term pushes the corresponding, i.e., augmented, instances toward each other in the RKHS and pulls away the non-corresponding instances away from each other in the RKHS.
This enforces consistency across augmentations in RKHS.

\subsection{Overall Kernel VICReg Loss}

Combining all three regularization terms, the final Kernel VICReg loss is given by:
\begin{equation}\label{equation_loss_kernel_vicreg}
\begin{aligned}
\mathcal{L}_{\text{Kernel-VICReg}} =\, &\alpha \mathcal{L}_{\text{inv}}(\bm{x}, \bm{x}') + \beta \big( \mathcal{L}_{\text{var}}(\bm{x}) + \mathcal{L}_{\text{var}}(\bm{x}') \big) \\
    &+ \zeta \big( \mathcal{L}_{\text{cov}}(\bm{x}) + \mathcal{L}_{\text{cov}}(\bm{x}') \big),
\end{aligned}
\end{equation}
where $\alpha, \beta, \zeta$ are hyperparameters controlling the contributions of invariance, variance, and covariance.
Note that the best values for hyperparameters $\alpha, \beta, \zeta, \gamma, \epsilon$ differ across the VICReg and Kernel VICReg. The best hyperparameters can be found depending on the dataset, as they are found in the original VICReg. 

By reformulating VICReg in RKHS, our method enables self-supervised learning in high-dimensional implicit feature spaces without explicit feature extraction, making it a powerful framework for non-linear representation learning.

\section{Discussions}

\subsection{Relation of Kernelized Variance Term with Kernel PCA}\label{section_relation_with_kernel_pca}

There is a close relation between the kernelized variance term in Kernel VICReg and kernel Principal Component Analysis (kernel PCA) \cite{scholkopf1997kernel}. 
In standard PCA, the eigenvalues of the covariance matrix $\bm{C} \in \mathbb{R}^{p \times p}$ are calculated using the following eigenvalue problem:
\begin{align}
\bm{C} \bm{u}_i = \eta_i \bm{u}_i,
\end{align}
where $\eta_i \in \mathbb{R}$ and $\bm{u}_i \in \mathbb{R}^p$ are the $i$-th eigenvalue and eigenvector of the covariance matrix $\bm{C}$, respectively. 

In kernel PCA, the eigenvalue problem of the double-centered kernel matrix\linebreak $\widehat{\bm{K}}(\bm{x}) \in \mathbb{R}^{b \times b}$ is considered: 
\begin{align}\label{sequation_kernel_pca_eigenvalue_problem}
\widehat{\bm{K}}(\bm{x}) \bm{v}_i = \lambda_i \bm{v}_i,
\end{align}
where $\lambda_i \in \mathbb{R}$ and $\bm{v}_i \in \mathbb{R}^b$ are the $i$-th eigenvalue and eigenvector of the double-centered kernel matrix $\widehat{\bm{K}}(\bm{x})$, respectively. 

Each eigenvector $\bm{v}_i$ gives a principal direction in the feature space $\phi(\bm{U}) \in \mathbb{R}^{t \times b}$. According to the representation theory, any function in RKHS lies in the span of all points in the RKHS \cite{mika1999fisher}:  
\begin{align}\label{equation_phi_u_Phi_V}
\phi(\bm{U}) = \sum_{i=1}^b \bm{\alpha}_i \phi(\bm{x}_i) = \bm{\Phi} \bm{A},
\end{align}
where $\bm{\Phi} \in \mathbb{R}^{t \times b}$ is defined in Equation (\ref{equation_Phi_matrix}) and $\bm{A} := [\bm{\alpha}_1, \dots, \bm{\alpha}_b]^\top \in \mathbb{R}^{b \times b}$ is the matrix of coefficients in the linear combination. 

On the one hand, the variance of the principal direction in the feature space is:
\begin{align*}
\textbf{Var}(\phi(\bm{U})) &= \frac{1}{b} \|(\bm{\Phi} \bm{H})^\top \bm{A}\|_2^2 \\
&=  \frac{1}{b} \textbf{tr}\big( ((\bm{\Phi} \bm{H})^\top \bm{A})^\top ((\bm{\Phi} \bm{H})^\top \bm{A}) \big) \\
&= \frac{1}{b} \textbf{tr}\big( \bm{A}^\top \bm{\Phi} \bm{\Phi}^\top \bm{A} \big) \\
&\overset{(\ref{equation_phi_u_Phi_V})}{=} \frac{1}{b} \textbf{tr}\big( \bm{A}^\top \underbrace{(\bm{\Phi} \bm{H})^\top (\bm{\Phi} \bm{H})}_{\widehat{\bm{K}}(\bm{x})} \underbrace{(\bm{\Phi} \bm{H})^\top (\bm{\Phi} \bm{H})}_{\widehat{\bm{K}}(\bm{x})} \bm{A} \big) \\
&= \frac{1}{b} \textbf{tr}\big( \bm{A}^\top \widehat{\bm{K}}(\bm{x})^2 \bm{A} \big).
\end{align*}
For one of the coefficient vectors, this equation becomes:
\begin{align}\label{equation_variance_phi_u}
\textbf{Var}(\phi(\bm{u}_i)) = \frac{1}{b} \bm{\alpha}_i^\top \widehat{\bm{K}}(\bm{x})^2 \bm{\alpha}_i,
\end{align}
where trace is dropped because the trace of a scalar is equal to itself. 

Assume that the coefficient $\bm{\alpha}_i$ is the $i$-th eigenvector of the double-centered kernel matrix. Thus, the variance becomes:
\begin{align}\label{equation_variance_phi_u_with_v}
\textbf{Var}(\phi(\bm{u}_i)) = \frac{1}{b} \bm{v}_i^\top \widehat{\bm{K}}(\bm{x})^2 \bm{v}_i.
\end{align}

Squaring the double-centered kernel matrix in Equation (\ref{sequation_kernel_pca_eigenvalue_problem}) gives:
\begin{align}\label{sequation_kernel_pca_eigenvalue_problem_squared}
\widehat{\bm{K}}(\bm{x})^2 \bm{v}_i = \lambda_i^2 \bm{v}_i.
\end{align}
Substituting Equation (\ref{sequation_kernel_pca_eigenvalue_problem_squared}) in Equation (\ref{equation_variance_phi_u_with_v}) provides:
\begin{align}
\textbf{Var}(\phi(\bm{u}_i)) &= \frac{1}{b} \bm{\alpha}_i^\top \lambda_i^2 \bm{\alpha}_i = \frac{1}{b} \lambda_i^2 \bm{\alpha}_i^\top \bm{\alpha}_i \nonumber \\
&= \frac{1}{b} \lambda_i^2 \|\bm{\alpha}_i\|_2^2. \label{equation_variance_phi_u_2}
\end{align}
The coefficient is usually normalized to have $\lambda_i \|\bm{\alpha}_i\|_2^2 = 1$ (so that $\|\phi(\bm{u}_i)\|_{\mathcal{H}} = 1$):
\begin{align}\label{equation_alpha}
\bm{\alpha}_i = \frac{\bm{v}_i}{\sqrt{\lambda_i}} \implies \| \bm{\alpha}_i \|_2^2 = \frac{1}{\lambda_i},
\end{align}
where the eigenvector $\bm{v}_i$ is assumed to be normalized to have unit length, i.e., $\|\bm{v}_i \|_2 = 1$.
According to Equation (\ref{equation_alpha}), Equation (\ref{equation_variance_phi_u_2}) becomes:
\begin{align}\label{equation_var_mu_over_b}
\textbf{Var}(\phi(\bm{u}_i)) &= \frac{\lambda_i}{b}.
\end{align}
This proves why $\lambda_i / b$ is used as variance in Equation (\ref{equation_kernelized_variance_term}).

On the other hand, left-multiplying Equation (\ref{sequation_kernel_pca_eigenvalue_problem}) by $\bm{v}_i^\top$ gives:
\begin{align*}
\bm{v}_i^\top \widehat{\bm{K}}(\bm{x}) \bm{v}_i = \bm{v}_i^\top \lambda_i \bm{v}_i = \lambda_i \bm{v}_i^\top \bm{v}_i = \lambda_i \| \bm{v}_i \|_2^2 \overset{(a)}{=} \lambda_i,
\end{align*}
where $(a)$ assumes that the eigenvector $\bm{v}_i$ is normalized to have unit length. 
Therefore, if $\|\bm{v}_i \|_2 = 1$, then:
\begin{align}\label{equation_mu_v_K_v}
\lambda_i = \bm{v}_i^\top \widehat{\bm{K}}(\bm{x}) \bm{v}_i.
\end{align}

Putting Equation (\ref{equation_mu_v_K_v}) in Equation (\ref{equation_var_mu_over_b}) provides:
\begin{align}
\textbf{Var}(\phi(\bm{u}_i)) = \frac{1}{b} \bm{v}_i^\top \widehat{\bm{K}}(\bm{x}) \bm{v}_i.
\end{align}

This analysis shows that the kernelized variance regularization in the proposed Kernel VICReg can be considered as kernel PCA. 
A similar analysis can be discussed to analyze the variance regularization term in VICReg as PCA. 




\subsection{Connection to HSIC and Independence}

The squared Hilbert--Schmidt norm of the RKHS covariance operator (used in our covariance loss) is closely related to the Hilbert--Schmidt Independence Criterion (HSIC), a well-established kernel-based dependence measure. In this view, our covariance loss can be interpreted as minimizing feature dependence in RKHS, encouraging the learning of diverse and disentangled features. This theoretical grounding strengthens the regularization effect of the Kernel VICReg loss beyond simple decorrelation.



\subsection{Kernel Choice as Inductive Bias}

Different kernels induce different geometric priors. For example, the RBF kernel emphasizes local smoothness, the Laplacian kernel allows sharper decision boundaries, and the rational quadratic interpolates between them. Our experiments reveal that no single kernel is optimal across all datasets; instead, performance depends on the match between the dataset structure and the kernel-induced geometry. This makes Kernel VICReg not only robust but also adaptable to task-specific data distributions. { A compatible extension to reduce sensitivity to a single kernel choice is to use kernel mixtures,
e.g., $k=\sum_m w_m k_m$ with $w_m\ge 0$ and $\sum_m w_m=1$, or more broadly multiple kernel learning (MKL).
This is orthogonal to our main contribution (lifting VICReg into RKHS), since the objective depends on Gram matrices and can directly operate on a mixture Gram matrix.}

{
\subsection{Comparison of Approaches in Kernel VICReg and Graph-Based Embedding}

There exist recent works on kernel-graph integration and spectral clustering 
\linebreak (e.g., \cite{berahmand2025comprehensive,zhu2021empirical}) analyzing how kernel methods interact with graph structure learning and spectral objectives. 
These approaches focus primarily on unsupervised clustering or graph-based embedding construction.
In contrast, our work does not construct or learn a graph structure; instead, we lift a non-contrastive self-supervised regularization objective into RKHS. The role of kernels here is not spectral clustering, but redefining variance and covariance operators in an implicit feature space. 
Our formulation in Kernel VICReg is therefore complementary rather than
competitive with kernel--graph hybrids.
}

{
\subsection{Comparison of Optimization Objective in Kernel VICReg and Variational Inference}

Although our formulation pulls the VICReg objective to an RKHS via kernel covariance operators, the resulting loss remains a deterministic, differentiable functional of the network parameters. Unlike variational self-supervised methods that introduce latent variables and evidence lower bounds (ELBO), our objective does not involve probabilistic latent modeling or variational inference. 

Concretely, the kernelized variance and covariance terms are computed through empirical covariance operators constructed from mini-batch embeddings. These operators depend smoothly on the encoder parameters, and gradients are obtained via automatic~\mbox{differentiation}.

Therefore, the optimization problem is a standard stochastic minimization of a deterministic loss (\ref{equation_loss_kernel_vicreg}) optimized using stochastic gradient descent. No variational bound, alternating optimization, or EM-style procedure is introduced by the RKHS lifting.
}

{
\subsection{Theoretical Properties of Kernel VICReg}

\subsubsection{Non-Collapse in RKHS}

\begin{prop}[Non-Collapse in RKHS]\label{proposition_non_collapse_rkhs}
Let $\widehat{\bm{K}}(\bm{x})$ denote the double-centered kernel matrix of a batch.
If the kernelized variance regularization enforces:
\begin{align}
\sqrt{\frac{\lambda_i}{b}} \geq \gamma > 0,
\quad \forall i \in \{1, \dots, b\},
\end{align}
where $\{\lambda_i\}_{i=1}^b$ are eigenvalues of $\widehat{\bm{K}}(\bm{x})$, then the covariance operator $\bm{C}_{\phi}(\bm{x})$ in RKHS is strictly positive definite on the span of the batch,
and representational collapse (i.e., rank-one embedding) is~prevented.
\end{prop}

\begin{proof}
From Equation (\ref{equation_C_proportional_K}), we have $\bm{C}_{\phi}(\bm{x}) \propto (1/b) \widehat{\bm{K}}(\bm{x})$. If all eigenvalues satisfy $\lambda_i > 0$,
then $\widehat{\bm{K}}$ is full rank on the batch span.
Thus the covariance operator has a strictly positive spectrum,
implying that no direction in RKHS has zero variance.
A collapsed representation corresponds to
$\operatorname{rank}(\widehat{\bm{K}}) = 1$,
which contradicts the enforced lower bound.
\end{proof}

\begin{Remark}
Proposition \ref{proposition_non_collapse_rkhs} demonstrates that Kernel VICReg enforces spectral spread in RKHS, while Euclidean VICReg only enforces coordinate-wise variance. This is a theoretical distinction between VICReg and Kernel KICReg.
\end{Remark}

\subsubsection{Nonlinear Variance Capture in RKHS}




\begin{Theorem}[Nonlinear Variance Capture in RKHS]\label{theorem_nonlinear_variance_capture}
Let $\mathcal{M} \subset \mathbb{R}^p$ be a compact nonlinear manifold.
Assume that $\mathcal{M}$ is not contained in any proper affine subspace 
of $\mathbb{R}^p$, but its nonlinear structure cannot be captured 
by second-order Euclidean statistics (i.e., PCA does not linearize $\mathcal{M}$).

Let $k$ be a universal kernel (e.g., Gaussian RBF or Laplacian) 
with feature map $\phi : \mathbb{R}^p \to \mathcal{H}$.
Then:
\begin{enumerate}
\item The feature map $\phi$ is injective on $\mathcal{M}$.
\item The image $\phi(\mathcal{M})$ lies in a linear subspace of $\mathcal{H}$ whose covariance operator encodes nonlinear structure of $\mathcal{M}$.
\item The eigenvalues of the centered kernel matrix correspond to nonlinear principal components of $\mathcal{M}$ (kernel PCA).
\item Therefore, enforcing lower bounds on eigenvalues of $\widehat{\bm{K}}(\bm{x})$ preserves nonlinear modes of variation that are invisible to Euclidean covariance regularization.
\end{enumerate}
\end{Theorem}

\begin{proof}
See Appendix \ref{section_appendix_proof_theorem_nonlinear_variance_capture} for proof. 
\end{proof}

\begin{Remark}
Theorem \ref{theorem_nonlinear_variance_capture} connects Kernel VICReg to kernel PCA theory, as was also discussed in Section \ref{section_relation_with_kernel_pca}.
This theorem does not claim that RKHS variance always 
strictly dominates Euclidean variance, 
but rather that for universal kernels, 
nonlinear structure becomes linearly representable in feature space, 
allowing spectral regularization to act on intrinsic manifold directions.
\end{Remark}

\subsubsection{Spectral Stability in Small Batches}




\begin{Theorem}[Spectral Stability of Centered Kernel Matrices]\label{theorem_spectral_stability}
Let $\{\bm{x}_i\}_{i=1}^b$ be i.i.d. samples drawn from a distribution $\mathcal{D}$.
Let $k$ be a bounded kernel satisfying:
\begin{align}\label{equation_kernel_bounded}
0 \le k(\bm{x},\bm{x}) \le \kappa^2, \quad \forall \bm{x}.
\end{align}
Let $\bm{K}_b$ denote the $b \times b$ Gram matrix,
and let $\widehat{\bm{K}}_b = \bm{H} \bm{K}_b \bm{H}$
be its double-centered version, where $\bm{H} = \bm{I}_b - (1/b) \mathbf{1}\mathbf{1}^\top$.
Let $\bm{\Sigma} := \mathbb{E}[\widehat{\bm{K}}_b]$ denote the population centered Gram operator
restricted to the sample span.
Then, for any $\delta \in (0,1)$,
with high probability at least $1-\delta$, we have:
\begin{align}\label{equation_bound}
\|\widehat{\bm{K}}_b - \bm{\Sigma}\|_2
\;\le\;
c \kappa^2
\left(
\sqrt{\frac{\log(2b/\delta)}{b}}
+
\frac{\log(2b/\delta)}{b}
\right),
\end{align}
for some universal constant $c>0$.
\end{Theorem}

\begin{proof}
See Appendix \ref{section_appendix_proof_theorem_spectral_stability} for proof. 
\end{proof}

The bound (\ref{equation_bound}) shows that eigenvalue estimates in RKHS concentrate at rate $\mathcal{O}(1/\sqrt{b})$, providing stability guarantees for small batch regimes.

\begin{Corollary}
Under the above conditions, each eigenvalue $\lambda_i$ of $\widehat{\bm{K}}_b$ concentrates around its population counterpart at rate $\mathcal{O}(1/\sqrt{b})$. Therefore, it provides stability guarantees for small-batch regimes where $b$ is not too large.
\end{Corollary}

}

{
\subsection{Scalability and Large-Scale Approximations}\label{section_scalability}
A primary concern in kernel-based methods is the computational complexity associated with the Gram (kernel) matrix. In Kernel VICReg, the construction of the kernel matrix $\bm{K} \in \mathbb{R}^{b \times b}$ and the eigenvalue decomposition required for the variance loss incur complexities of $\mathcal{O}(b^2)$ and $\mathcal{O}(b^3)$, respectively, where $b$ is the batch size, while modern hardware efficiently handles standard batch sizes (e.g., $b=256$ to $2048$), scaling to ``cognitive computing'' levels with massive datasets requires approximation strategies.

\subsubsection{Scalability by Nystr\"{o}m Method}

To address this, Kernel VICReg can be integrated with the Nystr\"{o}m method \cite{nystrom1930praktische,williams2000using}, which approximates the full kernel matrix using $m$ landmark points ($m \ll b$):
\begin{equation}
    \widetilde{\bm{K}} := \bm{K}_{b,m} \bm{K}_{m,m}^{-1} \bm{K}_{m,b},
\end{equation}
where $\bm{K}_{b,m}$ is the matrix of kernel evaluations between the batch samples and the landmarks. This reduces the complexity to $\mathcal{O}(bm^2)$.

\subsubsection{Scalability by Random Fourier Features}

Alternatively, Random Fourier Features (RFF) \cite{rahimi2007random} can be employed for shift-invariant kernels (e.g., RBF kernel). RFF maps the embeddings $\bm{z}$ to a low-dimensional randomized feature space $\Phi(\bm{z}) \in \mathbb{R}^D$ such that the kernel is approximated by a linear inner product:
\begin{equation}
\begin{aligned}
&k(\bm{z}_i, \bm{z}_j) \approx \Phi(\bm{z}_i)^\top \Phi(\bm{z}_j), \\
&\Phi(\bm{z}) = \sqrt{\frac{2}{D}} \left[ \cos(\bm{\omega}_1^\top \bm{z} + \beta_1), \dots, \cos(\bm{\omega}_D^\top \bm{z} + \beta_D) \right]^\top,
\end{aligned}
\end{equation}
where $\{\bm{\omega}_i\}_{i=1}^D$ are sampled from the kernel's spectral density. By utilizing RFF, the Kernel VICReg objective simplifies to a linear form in the $\Phi$-space, achieving $\mathcal{O}(bD)$ complexity. This ensures linear scalability relative to the batch size, making the framework suitable for large-scale distributed applications.
}

{
\subsection{Ethical Considerations and Potential Biases}
While Kernel VICReg provides a robust framework for nonlinear representation learning, it is essential to consider the ethical implications regarding data bias. Kernel methods are inherently sensitive to the distribution of the training data. In domains such as medical imaging or cognitive computing, if a specific group is underrepresented, the kernel matrix $\bm{K}$ may fail to capture the local geometry of that sub-population, potentially leading to ``feature exclusion'', where the RKHS mapping reinforces existing biases. 

Furthermore, the choice of kernel (e.g., the RBF kernel width $\sigma$) acts as a scale-dependent filter. If the data from minority groups exhibits different variance scales, a global kernel parameter might sub-optimally encode their features compared to the majority group. To mitigate these risks in sensitive applications, we suggest the use of adaptive kernels or group-fairness constraints within the variance-covariance terms, ensuring that the geometric embedding remains equitable across diverse demographic groups.
}

\section{Experimental Results}

We evaluate Kernel VICReg on a range of benchmark datasets to assess its ability to learn rich, non-linear representations in self-supervised settings. Our experiments span small-scale datasets (MNIST, CIFAR-10), mid-scale transfer learning (STL-10), and large-scale benchmarks (TinyImageNet, ImageNet100). For all experiments, we use a ResNet-18 backbone as the encoder, followed by a two-layer MLP projector. The model is trained using the Adam optimizer with an initial learning rate of $3 \times 10^{-4}$ batch size of 256, and cosine learning rate scheduling. Each dataset is augmented following standard protocols: random cropping, horizontal flipping, and color jitter for natural image datasets.

To investigate the effect of different kernels in the Reproducing Kernel Hilbert Space (RKHS), we implement Kernel VICReg using four kernels: linear, radial basis function (RBF), Laplacian, and rational quadratic (RQ). The kernel matrices are computed batch-wise, with double-centering applied to ensure zero-mean embeddings in RKHS. We evaluate the learned representations using linear probing, where a logistic regression classifier is trained atop frozen embeddings, and transfer learning, where encoders pretrained on CIFAR-10 are evaluated on STL-10.

\begin{table*}[t]
\caption{Linear 
evaluation on ImageNet100 and TinyImageNet with ResNet-18 backbone. The performances of baseline methods on TinyImageNet are adapted from \cite{liang2024multiple}.\\}
\label{tab:large_scale}
\begin{tabularx}{\linewidth}{LCC}
\toprule
\textbf{Model} & \textbf{ImageNet100 } & \textbf{TinyImageNet } \\
\midrule
SimCLR & 78.64 & 37.83 \\
SwAV & 74.36 & 35.39 \\
MoCo & 79.62 & 41.23 \\
BYOL & 80.88 & 36.31 \\
DINO & 75.41 & 35.77 \\
SimSiam & 78.80 & 27.96 \\
VICReg & 79.77 & Collapse \\
Barlow Twins & 80.63 & - \\
\midrule
Kernel VICReg---Linear (ours) & 77.34 & 37.48 \\
Kernel VICReg---RBF (ours) & 78.14 & 38.21 \\
Kernel VICReg---Laplacian (ours) & 79.92 & 40.12 \\
Kernel VICReg---RQ (ours) & 79.80 & 40.38 \\
\bottomrule
\end{tabularx}
\end{table*}

\subsection{Comparison with Baselines}

We compare Kernel VICReg against a diverse set of prominent self-supervised learning methods, including contrastive (SimCLR, MoCo), clustering-based (SwAV), and non-contrastive frameworks (BYOL, DINO, SimSiam, Barlow Twins, and VICReg). All baselines are implemented with the same ResNet-18 encoder to ensure fairness, and we reuse reported numbers or reproduce them when necessary using identical augmentation and optimization settings.

{
While the field of self-supervised learning (SSL) is rapidly evolving with newer non-contrastive architectures, the primary objective of this work is to provide a theoretical and methodological framework for lifting the VICReg objective into the Reproducing Kernel Hilbert Space (RKHS). Our evaluations focus on comparing Kernel VICReg against its direct Euclidean counterpart and established foundational SSL baselines (e.g., SimCLR, Barlow Twins, and VICReg) to isolate the impact of kernelization. By demonstrating consistent improvements over the original VICReg across multiple datasets, we establish the validity of the proposed kernelized variance-invariance-covariance constraints. 
}

Table~\ref{tab:large_scale} summarizes the top-1 linear evaluation accuracy on ImageNet100 and TinyImageNet. While VICReg exhibits competitive performance on ImageNet100, it collapses on TinyImageNet due to its sensitivity to small datasets with high intra-class variance. In contrast, Kernel VICReg remains stable across all settings, with the Laplacian and RQ kernels achieving the best performance, demonstrating the robustness of RKHS-based~\mbox{regularization}.

\begin{figure*}[t]
 \hspace{-7mm}   \includegraphics[width=0.49\textwidth]{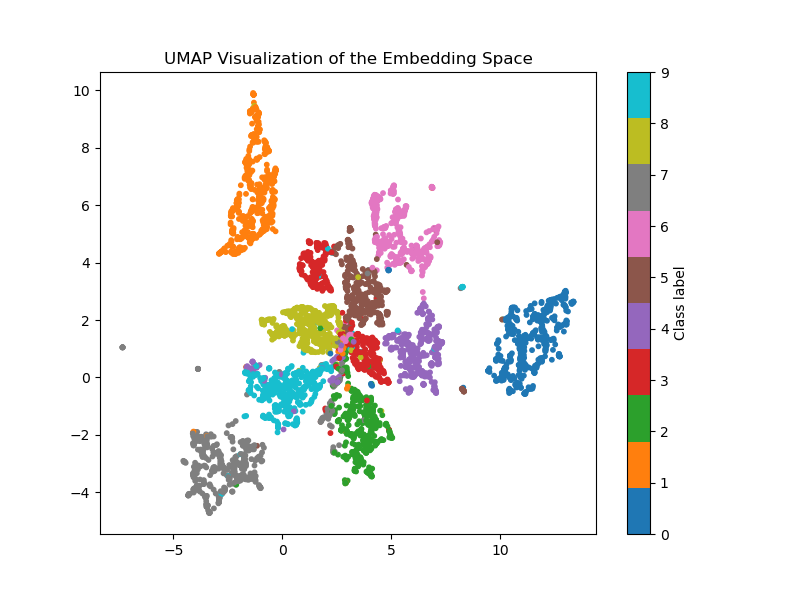}
    \includegraphics[width=0.49\textwidth]{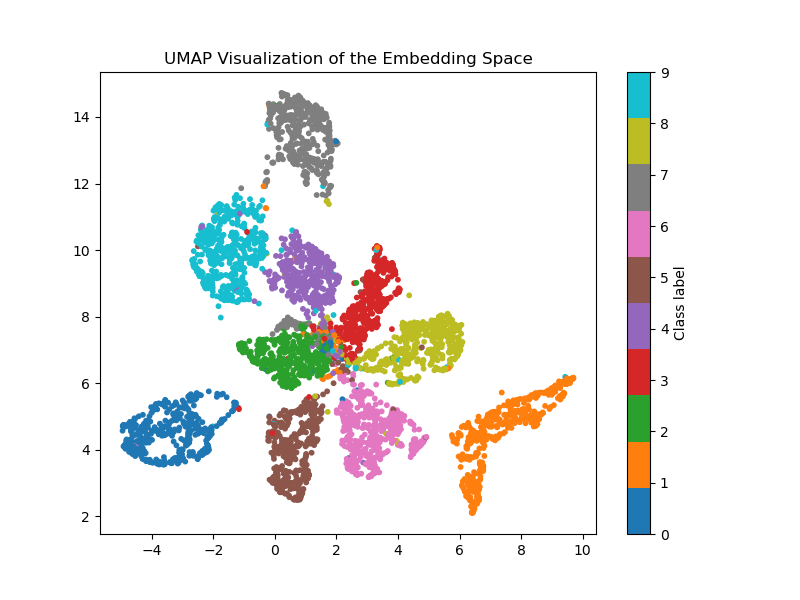}
 \centering   \includegraphics[width=0.49\textwidth]{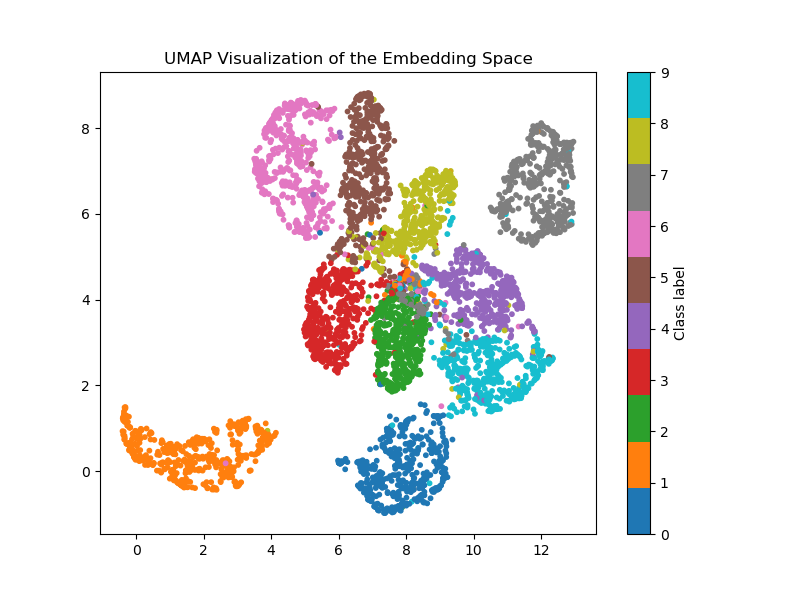}
    \caption{{UMAP 
projections (axes: UMAP-1 and UMAP-2) of MNIST embeddings from VICReg (\textbf{top left}), Kernel VICReg with linear kernel (\textbf{top right}), and Kernel VICReg with Laplacian kernel (\textbf{bottom}). Colors denote digit classes (label indices 0--9). The Laplacian kernel yields rounder, more isometric clusters, indicating improved class separability.}}
    \label{fig:umap}
\end{figure*}

Table~\ref{tab:small_scale} presents results on MNIST and CIFAR-10. Kernel VICReg consistently outperforms its Euclidean counterpart, particularly on MNIST, where the Laplacian kernel reaches $98.50\%$ accuracy. On CIFAR-10, the RQ kernel achieves the best performance at $86.18\%$, indicating that kernel choice adapts to data complexity.

\begin{table}[t]
\caption{Linear 
evaluation on MNIST and CIFAR-10 with ResNet-18 backbone.\\}
\label{tab:small_scale}
\begin{tabularx}{\linewidth}{lCC}
\toprule
\textbf{Model} & \textbf{MNIST} & \textbf{CIFAR-10} \\
\midrule
VICReg & 97.15 & 83.41 \\
Kernel VICReg---Linear & 98.33 & 86.08 \\
Kernel VICReg---RBF & 92.68 & 81.13 \\
Kernel VICReg---Laplacian & 98.50 
 & 84.56 \\
Kernel VICReg - RQ & 97.46 & 86.18 \\
\bottomrule
\end{tabularx}
\end{table}

Finally, Table~\ref{tab:transfer} reports transfer learning results on STL-10 using encoders pretrained on CIFAR-10. Kernel VICReg transfers better than VICReg, highlighting its generalization capabilities in low-label regimes.

{
Note that while several variations and incremental improvements to the VICReg architecture have been proposed since its inception in 2022, this study focuses on the foundational task of extending the core VICReg objective into the RKHS. By comparing our method directly against the standard Euclidean VICReg in Tables~\ref{tab:small_scale} and \ref{tab:transfer}, we isolate the performance gains attributable to the kernelized covariance and variance constraints. This controlled comparison is essential for validating the theoretical derivation of \mbox{Kernel~VICReg.} 
}

\begin{table}[t]
\caption{Transfer learning accuracy on STL-10. Embeddings are trained on CIFAR-10 (ResNet-18).\\}
\label{tab:transfer}
\begin{tabularx}{\linewidth}{lc}
\toprule
\textbf{Model} & \textbf{STL-10 Accuracy} \\
\midrule
VICReg & 69.82 \\
Kernel VICReg---Linear & 71.38 \\
Kernel VICReg---RBF  & 67.21 \\
Kernel VICReg---Laplacian & 71.09 \\
Kernel VICReg---RQ  & 72.34 \\
\bottomrule
\end{tabularx}
\end{table}




\subsection{Further Analysis and Insights}

To better understand how kernelization affects the structure of learned representations, we visualize the embedding spaces on the MNIST dataset using UMAP for three models: original VICReg, Kernel VICReg with a linear kernel, and Kernel VICReg with a Laplacian kernel (see Figure~\ref{fig:umap}). The UMAP plots reveal differences in cluster geometry across these methods. Representations from standard VICReg exhibit some class separation, but the clusters are elongated and lack compactness, suggesting anisotropic variance and potential feature collapse. The red cluster in VICReg is separated and is constructing two clusters; however, this is not the case for our method.

Kernel VICReg with a linear kernel improves upon this, producing tighter and more separated clusters, indicating that even without explicit nonlinearity in the kernel, RKHS-based decorrelation provides better structure. However, the most striking improvement appears with the Laplacian kernel: clusters become nearly circular and uniformly spaced, exhibiting strong isometry. This implies that the Laplacian-induced RKHS preserves pairwise relations and local structure more effectively, leading to embeddings with more consistent intra-class variance and improved inter-class margins.

{ These visualizations are qualitative; quantitative comparisons are provided by the linear-probe and transfer results in Tables~\ref{tab:large_scale} and \ref{tab:transfer}.} As shown earlier in Table~\ref{tab:large_scale}, Kernel VICReg maintains stable and competitive accuracy across both large-scale (ImageNet100) and small-scale (TinyImageNet) benchmarks. Notably, VICReg collapses on TinyImageNet, consistent with its known sensitivity to datasets exhibiting high intra-class variance or insufficient regularization. In contrast, kernelized versions (especially with Laplacian and rational quadratic kernels) perform robustly, demonstrating the benefits of nonlinear geometric alignment in RKHS.

\begin{figure}[t]
    \includegraphics[width=\linewidth]{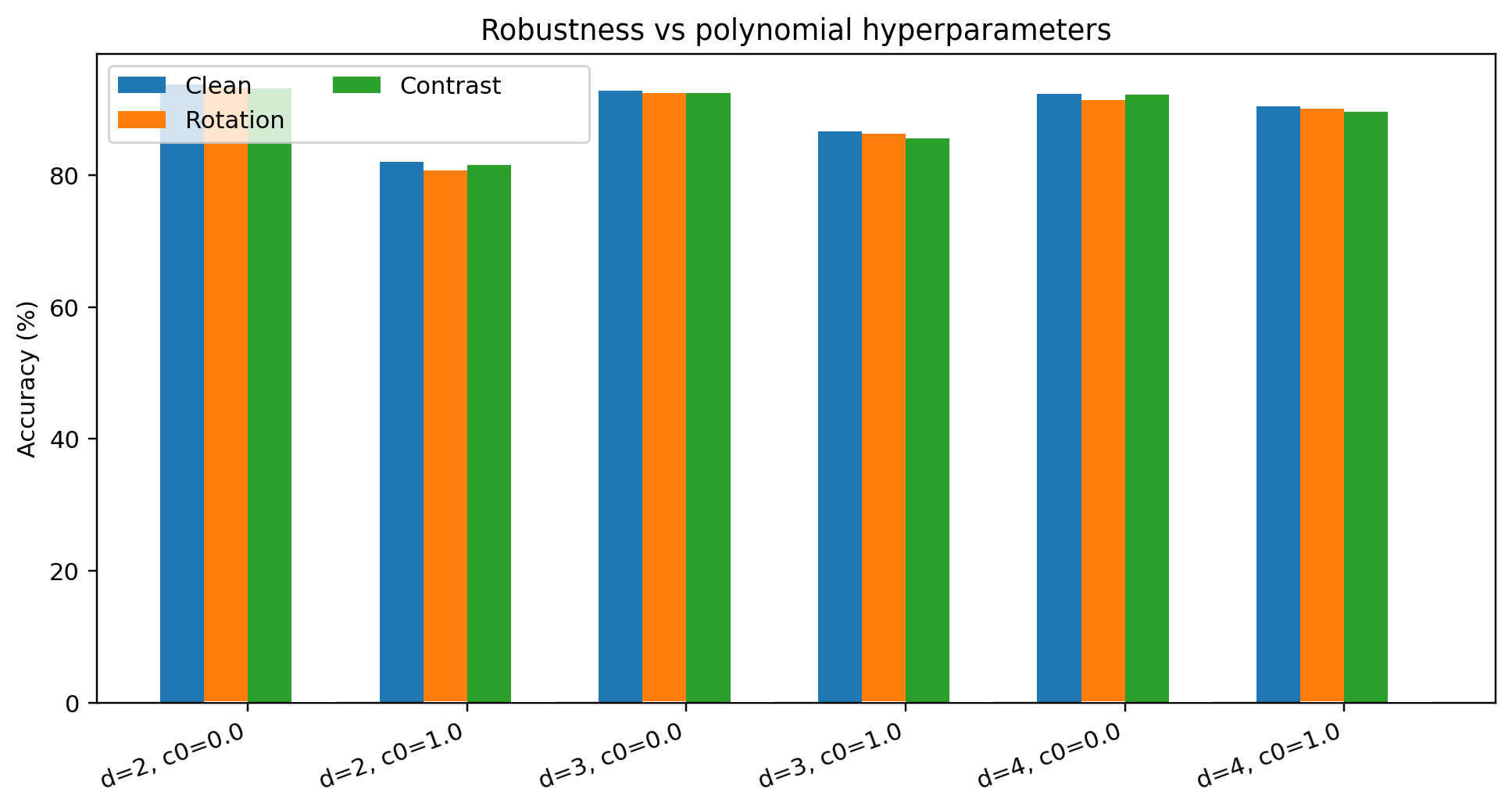}
    \caption{{Robustness 
of polynomial-kernel Kernel-VICReg across hyperparameter settings and distribution shifts on MNIST. Each grouped bar corresponds to one $(d,c_0)$ configuration, where $d$ is polynomial degree and $c_0$ is the additive constant in $k(x,y)=(\gamma x^\top y + c_0)^d$. Bars report linear-probe accuracy on clean data and under rotation and contrast shift. The spread across groups indicates strong hyperparameter sensitivity, with certain settings preserving clean/shifted performance.}}
    \label{fig:robustness_poly}
\end{figure}

\begin{figure}[t]
    \includegraphics[width=0.9\linewidth]{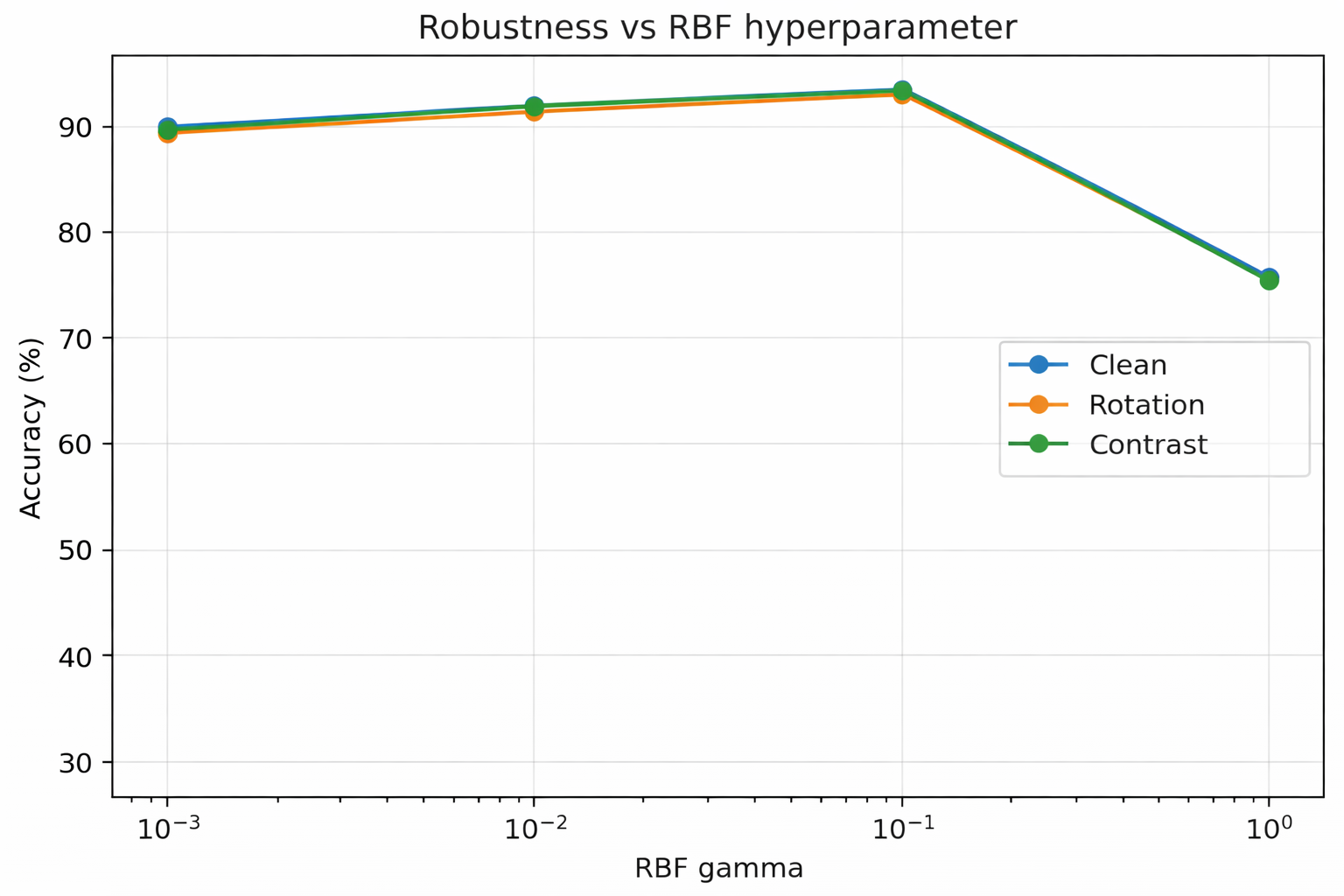}
    \caption{Robustness 
sensitivity of RBF-kernel Kernel-VICReg to the bandwidth parameter $\gamma$ under clean and shifted MNIST evaluation. Curves report linear-probe accuracy on clean data and under rotation and contrast shift. Performance is non-monotonic with respect to $\gamma$: moderate values yield better overall robustness, while overly large $\gamma$ leads to pronounced degradation, demonstrating the need for careful kernel hyperparameter tuning.}
    \label{fig:robustness_rbf_gamma}
\end{figure}

{We further analyze robustness sensitivity to kernel hyperparameters. Figure~\ref{fig:robustness_poly} shows that polynomial kernels can be highly sensitive to $(d,c_0)$ under distribution shifts, while Figure~\ref{fig:robustness_rbf_gamma} demonstrates that RBF performance is non-monotonic in the bandwidth parameter~$\gamma$.}

Furthermore, the results in Table~\ref{tab:transfer} highlight the generalization strength of kernel-based methods in transfer learning. Embeddings trained on CIFAR-10 and evaluated on STL-10 show that the RQ and Laplacian kernels outperform VICReg and even the linear kernel variant. These findings support the hypothesis that kernel-induced representations better capture underlying data manifolds, resulting in improved performance on downstream tasks with distributional shifts.

Overall, Kernel VICReg offers a principled extension of VICReg that gracefully incorporates nonlinearity through RKHS-based loss formulations. The improved cluster geometry, resilience to collapse, and higher transfer accuracy together suggest that kernelized self-supervision is a promising direction for representation learning beyond Euclidean~\mbox{limitations}.

\subsection{Implementation Details}

We implemented Kernel VICReg in PyTorch 2.6.0 
 with a modular design that allows kernel choice and parameter tuning through command-line flags. The training pipeline consists of three components: (i) a backbone encoder (either ResNet-18 or a simple CNN), (ii)~a multi-layer perceptron projector, and (iii) the proposed kernelized VICReg loss. Two augmented views are generated per sample following standard SSL protocols, and their embeddings are compared via the kernel-based losses.

\textbf{Kernels and Centering.} We implemented five kernels (linear, polynomial, radial basis function (RBF), Laplacian, and rational quadratic (RQ)) with automatic double-centering to operate in RKHS. For scale-sensitive kernels (RBF, Laplacian, and RQ), the bandwidth parameter $\gamma$ is adaptively estimated using the median heuristic unless explicitly specified.

\textbf{Kernel VICReg Loss.} The loss consists of three terms: (i) \emph{invariance}, computed as the trace distance between within-view and cross-view Gram matrices; (ii) \emph{variance}, computed from eigenvalues of the double-centered kernel matrix, penalizing directions with variance below a threshold; and (iii) \emph{covariance}, computed as the squared Hilbert--Schmidt norm of the covariance operator, equivalent to the sum of squared off-diagonal entries in the kernel covariance. The implementation returns all three contributions separately, allowing monitoring of invariance, variance, and covariance during training.

\textbf{Training Setup.} 
We conducted experiments on MNIST, CIFAR-10, STL-10, TinyImageNet, and ImageNet100. Unless otherwise noted, the encoder was a ResNet-18 backbone (with adjusted stem for CIFAR) followed by a 3-layer MLP projector with hidden dimension 1024. Models were trained using the Adam optimizer with a learning rate of $10^{-3}$, a batch size of 512, and a cosine learning rate schedule. Augmentations followed standard SSL protocols: random crop, color jitter, blur, and horizontal flip for natural images; affine transforms for~MNIST.

{ \textbf{Hyperparameters.} 
Table~\ref{table:coef} reports the best-performing $(\alpha,\beta,\zeta)$ found under our limited search for each dataset/kernel. 
We include these values to document the observed dataset dependence rather than to claim a single universally optimal setting. 
This dependence is expected because the kernel acts as an inductive bias that changes the geometry of the objective in RKHS.}

\begin{table*}[h]
\caption{Best-performing $(\alpha, \beta, \zeta)$ coefficients for Kernel VICReg across datasets.\\}
\begin{tabularx}{\textwidth}{lCCCCCCCCCCCC}
\toprule
\multirow{2}{*}{\vspace{-6pt}\centering\textbf{Kernel}} 
& \multicolumn{3}{c}{\textbf{MNIST}} 
& \multicolumn{3}{c}{\textbf{CIFAR-10}} 
& \multicolumn{3}{c}{\textbf{STL-10}} 
& \multicolumn{3}{c}{\textbf{TinyImageNet}} \\
\cmidrule{2-13}
& \boldmath{$\alpha$} & \boldmath{$\beta$} & \boldmath{$\zeta$} 
& \boldmath{$\alpha$} & \boldmath{$\beta$} &\boldmath{ $\zeta$} 
& \boldmath{$\alpha$} & \boldmath{$\beta$} & \boldmath{$\zeta$} 
& \boldmath{$\alpha$} & \boldmath{$\beta$} & \boldmath{$\zeta$} \\
\midrule
Linear     & 0.5  & 2  &  3 & 0.5  & 1  & 2  &  0.5 &  1 & 2  &  0.1 & 2  & 0.1  \\
RBF        & 0.5  & 2  & 2.5  & 0.5  & 1  & 3  &  0.5 &  1 & 2  &  0.1 & 2  & 0.2  \\
Laplacian   & 0.5  & 2  & 3   & 0.2   &1   & 2  &  0.5 & 1  &  2 & 0.1  & 2  & 0.1  \\
RQ         &  0.5 & 2  & 3  & 1.9  & 1  & 8  &  1.6 & 1  & 7  &  0.2 & 2  & 0.3  \\
\bottomrule
\end{tabularx}\label{table:coef}
\end{table*}

{ \textbf{Practical kernel-selection heuristics and reduced tuning.}
As a concise rule of thumb, Laplacian typically favors sharper/local structure (edge- and texture-dominated data), 
RBF favors smoother geometry and can be more forgiving under higher noise, 
and RQ is a practical middle ground when both local and global (multi-scale) structure matter. 
To reduce tuning burden in practice, one can fix $\alpha$ to a standard VICReg-scale value and perform small one-dimensional sweeps over $\beta$ (to prevent spectral collapse relative to $\gamma$) and $\zeta$ (to discourage redundancy), rather than jointly grid-searching all $(\alpha,\beta,$ and $\zeta)$.}

\textbf{Evaluation.} 
Representation quality was measured via \emph{linear probing}: embeddings from the frozen encoder were extracted, and a linear classifier was trained with SGD for 100~epochs. In addition, we visualized embedding geometry using UMAP and tracked top eigenvalue dynamics of the centered kernel matrix across epochs to study variance~\mbox{C}.

\begin{figure}[t]
    \includegraphics[width=0.9\linewidth]{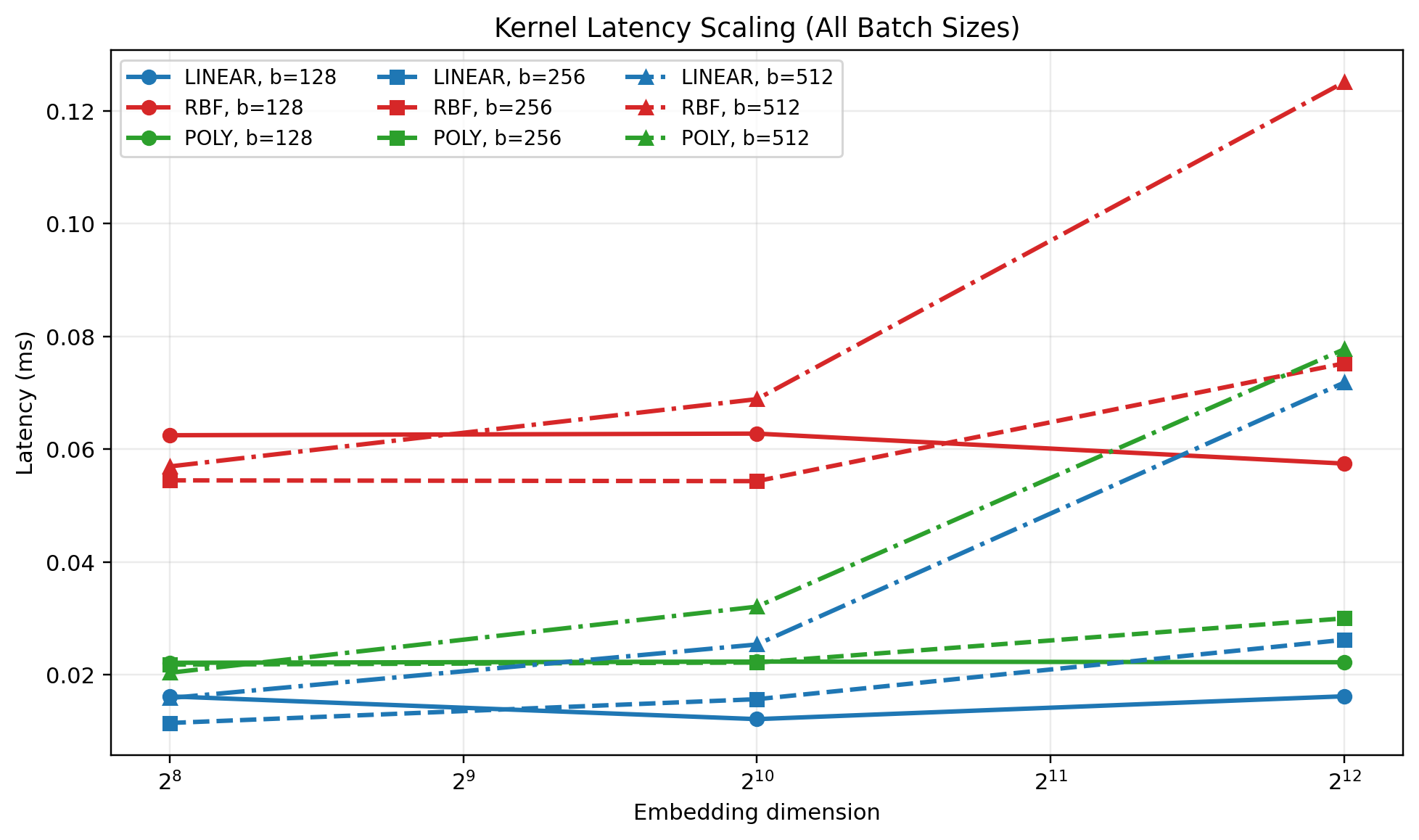}
    \caption{{Kernel latency scaling across embedding dimensions. Each curve corresponds to a kernel type (Linear, RBF, Polynomial) at a fixed batch size, and line style differentiates batch size while color differentiates kernel family. Latency increases with both embedding dimension and batch size, with RBF consistently incurring the highest compute cost and linear kernel remaining the fastest in most regimes. This figure quantifies the computational overhead of kernel choice under high-dimensional~settings.}}
    \label{fig:kernel_latency_scaling}
\end{figure}

\begin{figure}[t]
    \includegraphics[width=0.9\linewidth]{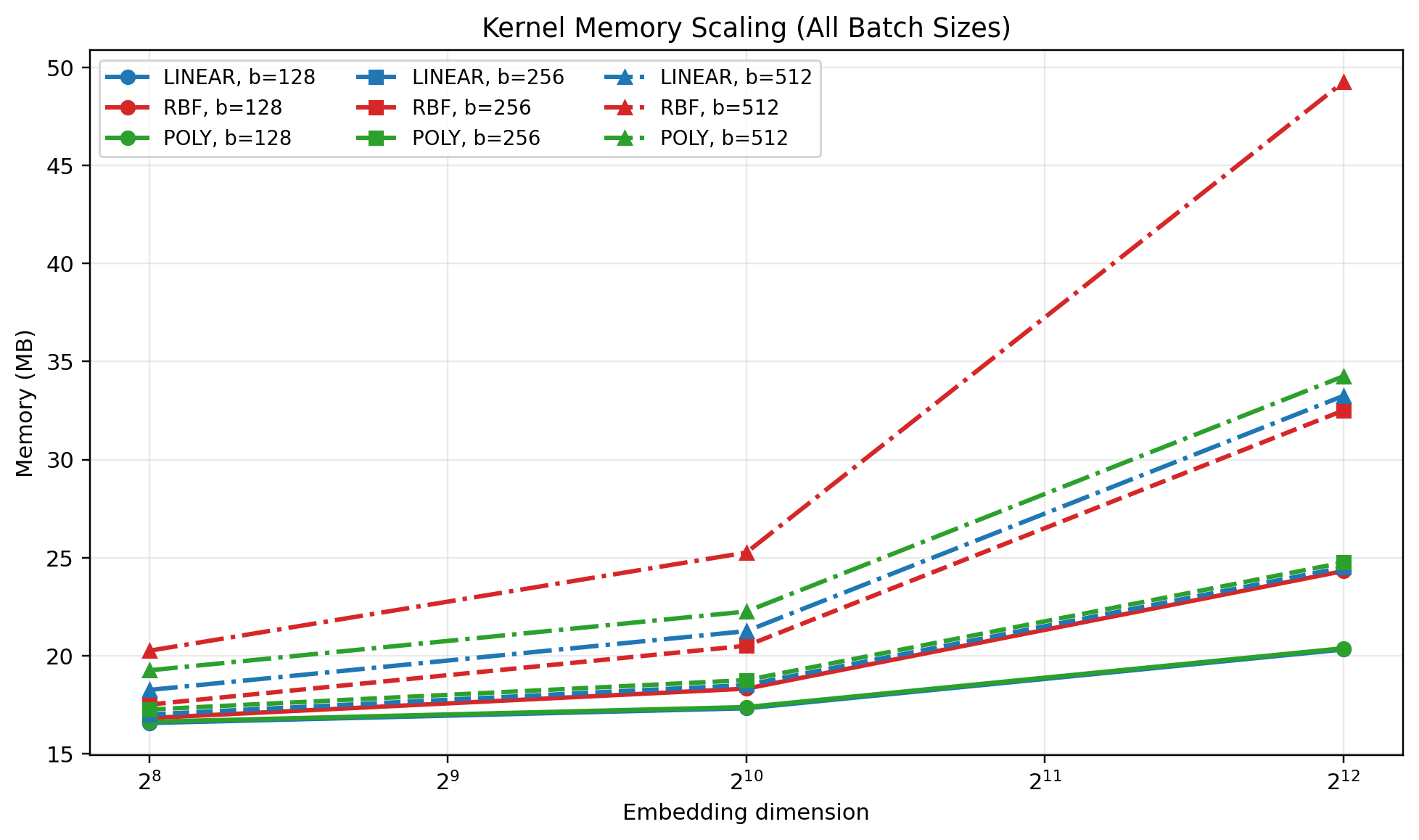}
    \caption{{ Kernel memory scaling across embedding dimensions. Color encodes kernel type (Linear, RBF, Polynomial), and line style encodes batch size. Memory usage increases with embedding dimension and batch size for all kernels, while RBF shows the highest memory footprint at larger dimensions, indicating its greater resource demand in high-dimensional regimes.}}
    \label{fig:kernel_memory_scaling}
\end{figure}

{
\subsection{Computational Complexity and Empirical Overhead}
While Kernel VICReg introduces nonlinear mapping via the RKHS, it is important to quantify the overhead relative to the standard Euclidean VICReg. Let $b$ denote the batch size and $p$ the dimensionality of the embeddings. Standard VICReg computes a covariance matrix in $O(bp^2 + p^3)$, whereas Kernel VICReg computes the Gram matrix and its eigenvalue decomposition in $O(b^2 p + b^3)$.
Figures~\ref{fig:kernel_latency_scaling} and~\ref{fig:kernel_memory_scaling} further quantify how kernel choice scales in latency and memory as embedding dimension and batch size increase.
}
{
The results indicate that for standard SSL batch sizes ($b \le 2048$), the overhead is marginal. As discussed in Section \ref{section_scalability}, for larger ``cognitive computing'' scales where $b$ increases significantly, the $O(b^3)$ bottleneck can be effectively mitigated using Nystr\"{o}m approximations or Random Fourier Features, which reduce the complexity back to a linear or quasi-linear relationship with batch size.
}


\section{Conclusions}

We introduced Kernel VICReg, a principled extension of VICReg that pulls self-supervised learning objectives from Euclidean space to Reproducing Kernel Hilbert Space (RKHS). By reformulating invariance, variance, and covariance terms in kernel space using double-centered kernel matrices and Hilbert--Schmidt norms, our method captures complex nonlinear structures without requiring explicit feature mappings.

Our empirical results across diverse datasets demonstrate the robustness and effectiveness of this kernelized formulation. Kernel VICReg outperforms its Euclidean counterpart, particularly in challenging regimes such as TinyImageNet, where standard VICReg collapses. Moreover, kernel-induced representations exhibit superior generalization in transfer learning tasks, as evidenced by improvements on STL-10. Visualization through UMAP further reveals that kernelization promotes more compact, isometric cluster structures, especially under the Laplacian kernel.

These findings suggest that kernel methods offer a natural and powerful means to enhance self-supervised learning while our work focuses on VICReg, the framework readily extends to other SSL objectives such as Barlow Twins and SimCLR, opening promising avenues for future research in kernelized SSL. This study contributes to bridging classical kernel theory and modern representation learning by showing that the integration of RKHS structure meaningfully improves both stability and expressiveness in self-supervised~\mbox{models.}

\vspace{6pt}
\textbf{Author contributions:} Conceptualization, M.H.S., B.G., and P.F.; methodology, M.H.S. and B.G.; software, M.H.S.; validation, M.H.S. and B.G.; formal analysis, M.H.S. and B.G.; investigation, M.H.S. and B.G.; writing---original draft preparation, M.H.S. and B.G.; writing---review and editing,  M.H.S., B.G., S.M., and P.F.; visualization, M.H.S.; supervision, P.F.; project administration, S.M.; All authors have read and agreed to the published version of the manuscript.

\textbf{Funding:} 
This work was supported by the Natural Sciences and Engineering Research Council of Canada (NSERC) through the Discovery Grants Program.



\textbf{Data availability:} The original data presented in the study are openly available in publicly accessible repositories: MNIST 
 (\url{http://yann.lecun.com/exdb/mnist/} (accessed on 01-08-2025)), CIFAR-10 (\url{https://www.cs.toronto.edu/~kriz/cifar.html} (accessed on 01-08-2025)), STL-10 (\url{https://cs.stanford.edu/~acoates/stl10/} (accessed on 01-08-2025)), TinyImageNet (\url{https://www.kaggle.com/c/tiny-imagenet} (accessed on 01-08-2025)), and ImageNet (\url{https://www.image-net.org/} (accessed on 01-08-2025)). ImageNet100 is a subset of ImageNet constructed for evaluation in this study; no new data were~created.

\section*{Appendices}

\appendix
\section[\appendixname~\thesection]{Proof for Theorem \ref{theorem_nonlinear_variance_capture}}\label{section_appendix_proof_theorem_nonlinear_variance_capture}

{

\textbf{Step 1: Injectivity of Universal Kernels.}

A kernel $k$ on a compact domain is universal if its RKHS is dense 
in $C(\mathcal{M})$ with respect to the supremum norm.
For universal kernels such as the Gaussian RBF and Laplacian kernels, 
the associated feature map $\phi$ is injective; that is:
\begin{align*}
\bm{x} \neq \bm{y} \implies \phi(\bm{x}) \neq \phi(\bm{y}).
\end{align*}
Hence, the embedding $\phi : \mathcal{M} \to \mathcal{H}$ 
is one-to-one and preserves all information about the~\mbox{manifold}.

\textbf{Step 2: Linearization of Nonlinear Structure.}

Although $\mathcal{M}$ may be nonlinear in $\mathbb{R}^p$, its image $\phi(\mathcal{M})$ lies in a (possibly infinite-dimensional)
Hilbert space $\mathcal{H}$, where linear combinations are permitted.
Kernel PCA theory \cite{scholkopf1998nonlinear} shows that principal components in RKHS correspond to eigenfunctions of the 
covariance operator:
\begin{align*}
\bm{C}_{\phi}(\bm{x}) = \frac{1}{b} \sum_{i=1}^b 
(\phi(\bm{x}_i) - \bar{\phi})(\phi(\bm{x}_i) - \bar{\phi})^\top.
\end{align*}
The eigenvalue problem in $\mathcal{H}$ reduces to the finite-dimensional 
eigenvalue problem of the centered kernel matrix:
\begin{align*}
\widehat{\bm{K}}(\bm{x}) \bm{v}_i = \lambda_i \bm{v}_i.
\end{align*}
Thus, eigenvalues $\{\lambda_i\}$ represent variances 
along nonlinear principal directions in feature~\mbox{space}.

\textbf{Step 3: Failure of Euclidean Covariance.}

In Euclidean space, covariance captures only second-order linear variation:
\begin{align*}
\bm{C}_{\text{Euc}} = \frac{1}{b} \bm{Z}^\top \bm{H} \bm{Z}.
\end{align*}
If $\mathcal{M}$ exhibits curvature or nonlinear embedding 
(e.g., Swiss roll or concentric circles), 
Euclidean PCA cannot flatten the manifold; 
variance is spread across multiple coordinates 
without revealing intrinsic nonlinear directions.
Thus, Euclidean variance regularization may fail to preserve intrinsic manifold modes.

\textbf{Step 4: RKHS Variance Captures Nonlinear Modes.}

In contrast, kernel PCA diagonalizes the covariance operator in $\mathcal{H}$.
Since universal kernels generate feature spaces dense in $C(\mathcal{M})$, 
nonlinear structure becomes linearly separable in $\mathcal{H}$.
Therefore, eigenvalues $\{\lambda_i\}$ of $\widehat{\bm{K}}(\bm{x})$ 
quantify nonlinear principal variances of $\mathcal{M}$.
Enforcing $\lambda_i / b \ge \gamma^2 > 0$ ensures preservation of nonlinear modes of variation.

\textbf{Conclusion.}

Because RKHS covariance eigenvalues correspond to nonlinear 
principal components, kernelized variance regularization 
preserves nonlinear structure that Euclidean covariance cannot capture.
}

{
\section{Proof for Theorem \ref{theorem_spectral_stability}}\label{section_appendix_proof_theorem_spectral_stability}

\textbf{Step 1: Decomposition of the Gram Matrix.}

Define feature map $\phi(\bm{x})$ in RKHS $\mathcal{H}$.
The (uncentered) Gram matrix satisfies:
\begin{align*}
\bm{K}_b[i,j] = \langle \phi(\bm{x}_i), \phi(\bm{x}_j) \rangle.
\end{align*}
Let $\bm{\Phi} = [\bm{\phi}(\bm{x}_1), \dots, \bm{\phi}(\bm{x}_b)]$.
Then, we have $\bm{K}_b = \bm{\Phi}^\top \bm{\Phi}$.
The centered Gram matrix is:
\begin{align*}
\widehat{\bm{K}}_b = \bm{H} \bm{\Phi}^\top \bm{\Phi} \bm{H}
= (\bm{\Phi} \bm{H})^\top (\bm{\Phi} \bm{H}).
\end{align*}
Thus, $\widehat{\bm{K}}_b$ is the finite-sample covariance operator expressed in kernel coordinates.

\textbf{Step 2: Operator Representation.}

Define centered feature vectors:
\begin{align*}
\widetilde{\phi}(\bm{x}_i)
=
\phi(\bm{x}_i) - \bm{\mu},
\quad
\bm{\mu} = \mathbb{E}[\phi(x)].
\end{align*}
The population covariance operator in RKHS is $\bm{C}(\bm{x}) := \mathbb{E}[\widetilde{\phi}(x) \otimes \widetilde{\phi}(x)]$.
The empirical covariance operator $\bm{C}_b(\bm{x})$ is $(1/b)
\sum_{i=1}^b \widetilde{\phi}(x_i) \otimes \widetilde{\phi}(x_i)$, as also stated in Equation (\ref{equation_covariance_in_RKHS_matrix_form}).
Moreover, according to the definition in the theorem, we have $\bm{\Sigma} := \mathbb{E}[\widehat{\bm{K}}_b]$. 
Thus:
\begin{align*}
&\widehat{\bm{K}}_b = b \bm{C}_b, \\
&\bm{\Sigma} = b \bm{C}.
\end{align*}
Thus, the bounding $\|\widehat{\bm{K}}_b - \bm{\Sigma}\|_2$
reduces to bounding operator deviation $\|\bm{C}_b - \bm{C}\|_2$.

\textbf{Step 3: Boundedness.}

According to Equation (\ref{equation_kernel_bounded}), the kernel is bounded:
\begin{align*}
\|\phi(\bm{x})\|_\mathcal{H}^2
=
k(\bm{x},\bm{x})
\le \kappa^2,
\end{align*}
therefore,
\begin{align*}
\|\widetilde{\phi}(\bm{x})\|_\mathcal{H}
\le 2\kappa.
\end{align*}
Hence, each rank-one operator,
\begin{align*}
\bm{X}_i
:=
\widetilde{\phi}(\bm{x}_i) \otimes \widetilde{\phi}(\bm{x}_i),
\end{align*}
satisfies
\begin{align*}
\|\bm{X}_i\|_2
=
\|\widetilde{\phi}(\bm{x}_i)\|^2
\le 4\kappa^2.
\end{align*}

\textbf{Step 4: Apply Matrix Bernstein Inequality.}

The operators:
\begin{align*}
\bm{Y}_i = \bm{X}_i - \mathbb{E}[\bm{X}_i],
\end{align*}
are independent mean-zero self-adjoint operators.
Matrix Bernstein inequality \cite{tropp2012user} states that for such operators:
\begin{align*}
\Pr\!\left(
\left\|
\frac{1}{b}
\sum_{i=1}^b \bm{Y}_i
\right\|_2
\ge t
\right)
\le
2b
\exp
\left(
\frac{-b t^2/2}
{\sigma^2 + \bm{R} t/3}
\right),
\end{align*}
where
\begin{align*}
\bm{R} = \max_i \|\bm{Y}_i\|_2 \le 8\kappa^2,
\end{align*}
and variance proxy
\begin{align*}
\sigma^2
\le
\frac{1}{b}
\sum_{i=1}^b
\mathbb{E}
[\|\bm{Y}_i\|_2^2]
\le
4 \kappa^4.
\end{align*}
Solving the inequality for probability $\delta$
gives:
\begin{align*}
\|\bm{C}_b - \bm{C}\|_2
\le
c
\kappa^2
\left(
\sqrt{\frac{\log(2b/\delta)}{b}}
+
\frac{\log(2b/\delta)}{b}
\right).
\end{align*}

\textbf{Step 5: Transfer to Kernel Matrix.}

According to step 2 of the proof, the bounding $\|\widehat{\bm{K}}_b - \bm{\Sigma}\|_2$
reduces to bounding operator deviation $\|\bm{C}_b - \bm{C}\|_2$. Therefore, we obtain the same deviation bound for $\|\widehat{\bm{K}}_b - \bm{\Sigma}\|_2$.
}

\bibliography{references}
\bibliographystyle{icml2025}

\end{document}